# INTERLEAVING SYNTAX AND SEMANTICS IN AN EFFICIENT BOTTOM-UP PARSER[*]

John Dowding, Robert Moore, François Andry[†], and Douglas Moran
SRI International
333 Ravenswood Avenue
Menlo Park, CA 94025
{dowding,bmoore,andry,moran}@ai.sri.com [‡]

## Abstract

We describe an efficient bottom-up parser that interleaves syntactic and semantic structure building. Two techniques are presented for reducing search by reducing local ambiguity: Limited left-context constraints are used to reduce local syntactic ambiguity, and deferred sortal-constraint application is used to reduce local semantic ambiguity. We experimentally evaluate these techniques, and show dramatic reductions in both number of chart edges and total parsing time. The robust processing capabilities of the parser are demonstrated in its use in improving the accuracy of a speech recognizer.

## INTRODUCTION

The parsing problem is typically framed as a recognition problem: Given a grammar and a word string, determine if the word string is a member of the language described by the grammar. For some applications, notably robust natural-language processing and spoken-language understanding, this is insufficient, since many utterances will not be accepted by the grammar, because of nonstandard language, inadequate grammatical coverage, or errors made in speech recognition. In these cases, it is still desirable to determine what well-formed phrases occurred in the word string, even when the entire string is not recognized. The goal of the parser described here is to construct a chart, as efficiently as possible, that contains all the syntactically well-formed semantically meaningful phrases that occur in the word string.

The most efficient practical context-free parsers (Earley, 1970; Graham, Harrison, and Ruzzo, 1980) are left-corner parsers, which gain efficiency by their ability to constrain the search to find only phrases that might contribute to a sentence that starts at the left edge of the string being parsed. These strong left-context syntactic constraints can prevent the parser from finding some phrases that are well-formed, however. This is a problem for us that is avoided by bottom-up parsers (Kasami, 1965; Younger, 1967), but at the expense of creating many more edges, which can lead to dramatic increases in parse time. Since our goal is to find only the phrases that are semantically meaningful as well as syntactically well-formed, we also need to compute semantic constraints for every syntactic phrase we construct. This requires making finer distinctions than syntax-only parsing, which can introduce additional ambiguity, multiplying the number of distinct phrases found and increasing parse time.

We describe two special techniques for speeding up bottom-up parsing by reducing local ambiguity without sacrificing completeness. One technique, "limited left-context checking," reduces local syntactic ambiguity; the other, "deferred sortal-constraint application," reduces local semantic ambiguity. Both techniques are applied to unification-based grammars. We analyze the performance of these techniques on a 194-utterance subset of the ARPA ATIS corpus (MADCOW, 1992), using a broad-coverage grammar of English. Finally, we present results using the output of the parser to improve the accuracy of a speech recognizer in a way that takes advantage of our ability to find all syntactically well-formed semantically meaningful phrases.

---

[*]This research was supported by the Advanced Research Projects Agency under Contract ONR N00014-90-C-0085 with the Office of Naval Research. The views and conclusions contained in this document are those of the authors and should not be interpreted as necessarily representing the official policies, either expressed or implied, of the Advanced Research Projects Agency of the U.S. Government.

[†]Current address: CAP GEMINI Innovation, 86-90 Rue Thiers, 92513-Boulogne Billancourt, France, andry@capsogeti.fr.

[‡]To appear in 'Proceedings of the $32^{nd}$ Annual Meeting of the Association for Computational Linguistics,' June 1994.

# SYNTACTIC PARSING

The parsing algorithm described here is implemented in the Gemini spoken-language understanding system (Dowding et al., 1993), which features a broad-coverage unification-based grammar of English, with independent syntactic, semantic and lexical components, in the style of the SRI Core Language Engine (Alshawi, 1992). Although we describe the syntactic parsing algorithm as though it were parsing purely context-free grammars, the ideas extend in a natural way to unification-based grammar parsing. While the chart for a context-free grammar contains edges labeled by atomic nonterminal symbols, the chart for a unification-based grammar contains edges labeled by complex feature-structure nonterminals. For efficiency, we maintain edges in the chart in only their most general form—new edges are added to the chart only if they are more general than existing edges, and we delete existing edges that are less general than the new edge. Like the Core Language Engine, we use a technique called packing to prevent local ambiguity from multiplying out into distinct edges at higher levels in the tree. Packing is implemented by collapsing phrasal analyses that share the same parent nonterminal and using only the parent for further processing.

## Limited Left-Context Checking

The motivation behind limited left-context checking is the observation that most of the phrases found by a pure bottom-up parser using our unification grammar contain syntactic gaps not licensed by any possible gap filler. In a pure bottom-up parser, syntactic gaps must be hypothesized between every pair of words and lead to many spurious phrases being built. Earlier work (Moore and Dowding, 1991) showed that over 80% of the edges built by a bottom-up parser using our grammar were in this class. Since these phrases are semantically incomplete, they are of no interest if they cannot be tied to a gap filler, even in the robust processing applications we are concerned with. Our approach is to use left-context checking in a limited way to restrict the construction of only this class of phrases.

We partition the set of grammatical categories in our grammar into two groups, context-independent and context-dependent. Context-independent phrases will be always be constructed bottom-up whenever possible. Context-dependent phrases will only be constructed if they are predicted by previously constructed phrases to the left. For our purposes, the set of context-dependent phrases are those that contain a syntactic gap with no gap filler, and the context-independent set is everything else. Note, however, that there is no constraint on the algorithm that forces this. If every grammatical category is context-dependent, then this algorithm reduces to a left-corner parser, and if every category is context-independent, then this algorithm reduces to a pure bottom-up parser. One caveat is that for the algorithm to work correctly, the set of context-dependent categories must be closed under the possible-left-corner-of relation.

The question remains of how to produce predictions for only those phrases in the context-dependent set. As in Earley's algorithm, predictions are implemented as dotted grammar rules. Unlike Earley's algorithm, however, predictions are used only to license the construction of context-dependent categories. Predictions are not created for context-independent categories, and they are not used in a completion phase to find new reductions.

Predictions deriving from rules that create context-dependent categories must themselves be predicted. Thus, predictions are also divided into context-independent and context-dependent. A context-independent prediction will always be added to the chart after the first child on the right-hand side has been found. A context-dependent prediction will only be added to the chart when the first child on the right-hand side has been found, and the head of the rule has been previously predicted or is a possible left corner of a category that has been previously predicted. Tables containing the possible context-dependent and context-independent predictions are constructed at compile time.

An outline of the parser algorithm is given in Figure 1. The algorithm is basically an all-paths, left-to-right, bottom-up parser, with the modifications that (1) the edge resulting from a reduction is added to the chart only if it is either a context-independent phrase or is predicted, and (2) predictions are added at each point in the input for the context-dependent phrases that are licensed at that point. Some details of the parser have been omitted, particularly those related to parsing unification-based grammars that do not arise when parsing context-free grammars. In addition, the parser maintains a skeletal copy of the chart in which edges are labeled only by the nonterminal symbols contained in their context-free backbone, which gives us more efficient indexing of the full grammar rules. Other optimizations include using one-word look-ahead before adding new predictions, and using restrictors (Shieber, 1985) to increase the generality of the predictions.

For grammar with start symbol $\Sigma$, phrase structure rules $P$, lexicon $L$, context-independent categories $CI$, and context-dependent categories $CD$; and for word string $w = w_1...w_n$:

```
if Σ ∈ CD, predict(Σ, 0);
add_empty_categories(0);
for i from 1 to n do
  foreach C such that C → w_i ∈ L do
    add_edge_to_chart(C, i-1, i);
    make_new_predictions(C, i-1, i);
    find_new_reductions(C, i-1, i)
  end
  add_empty_categories(i);
end

sub find_new_reductions(B, j, k) {
  foreach A and α such that A → αB ∈ P do
    foreach i such that i = match(α, j) do
      if A ∈ CD and predicted(A,i) or A ∈ CI
        add_edge_to_chart(A, i, k);
        make_new_predictions(A, i, k);
        find_new_reductions(A, i, k);
    end
  end
}
sub add_empty_categories(i) {
  foreach A such that A → ε ∈ P do
    if A ∈ CD and predicted(A,i) or A ∈ CI
      add_edge_to_chart(A, i, i);
      make_new_predictions(A, i, i);
      find_new_reductions(A, i, i);
  end
}
sub make_new_predictions(A, i, j) {
  foreach Aβ ∈ Predictions[i] do
    predict(β, j)
  end
  foreach H → AαBβ ∈ P such that
    H ∈ CI and B ∈ CD and β ∈ CI* do
      predict(αB, j)
  end
  foreach H → AαBβ ∈ P such that
    H ∈ CD and B ∈ CD and β ∈ CI*
    and predicted(H, i) or
    H left-corner-of C and predicted(C, i) do
      predict(αB, j)
  end
}
```

Figure 1: Limited Left-Context Algorithm

| Variant | Edges | Preds | Secs |
|---|---|---|---|
| Bottom-Up | 1191 | 0 | 14.6 |
| Limited Left-Context | 203 | 25 | 1.0 |
| Left-Corner | 112 | 78 | 4.0 |

Table 1: Comparison of Syntax-Only Parsers

## Comparison with Other Parsers

Table 1 compares the average number of edges, average number of predictions, and average parse times[1] (in seconds) per utterance for the limited left-context parser with those for a variant equivalent to a bottom-up parser (when all categories are context independent) and for a variant equivalent to a left-corner parser (when all categories are context dependent). The tests were performed on a set of 194 utterances chosen at random from the ARPA ATIS corpus (MADCOW, 1992), using a broad-coverage syntactic grammar of English having 84% coverage of the test set.

The limited left-context parser can be thought of as at a midway point between the pure bottom-up parser and the left-corner parser, constructing a subset of the phrases found by the bottom-up parser, and a superset of the phrases found by the left-corner parser. Using limited left-context to constrain categories containing syntactic gaps reduces the number of phrases by more than a factor of 5 and is almost 15 times faster than the pure bottom-up parser. The limited left-context parser builds 81% more edges than the left-corner parser, but many fewer predictions. Somewhat surprisingly, this results in the limited left-context parser being 4 times faster than the left-corner parser. We conjecture that this is due to the fact that context-independent phrases are licensed by a static table that is quicker to check against than dynamic predictions. This results in a lower average time per edge for the limited left-context parser (0.005 seconds) than the left-corner parser (0.036 seconds). Some additional penalty may also have been incurred by not using dotted grammar rules to generate reductions, as in standard left-corner parsing algorithms.[2]

There are important differences between the technique for limited prediction in this parser, and other techniques for limited prediction such as Shieber's notion of restriction (Shieber, 1985) (which we also use). In methods such as Shieber's, predictions are weakened in ways that can result in an overall gain in efficiency, but predictions nevertheless must be dynamically generated for every phrase that is built bottom-up. In our method, *no* predictions need to be generated for the context-independent categories; from another point of view, context-independent categories are predicted statically, at compile time, for all points in the input, rather than dynamically at run time. Time is saved both because the predictions do not have to be generated at run time, and because the process of checking these static predictions is simpler.

---

[1] All parse times given in this paper were produced on a Sun SPARCstation 10/51, running Quintus Prolog version 3.1.4.

[2] Other than this, we do not believe that the bottom-up and left-corner algorithms we tested suffered from any unnecessary overheads from being implemented as special cases of our general algorithm, as we removed calls to subroutines that were unnecessary for those special cases.

In previous work (Moore and Dowding, 1991), we compared limited left-context checking to some other methods for dealing with empty categories in a bottom-up parser. Standard grammar transformation techniques (Hopcroft and Ullman, 1980) can be used to eliminate empty nonterminals. This approach is useful to eliminate some edges, but still allows edges that dominate empty categories to be created. We found that using this technique was faster than pure bottom-up parsing, but still significantly slower than limited left-context checking. A further refinement is to transform the grammar to eliminate both empty and nonbranching rules. In the case of our grammar, however, this resulted in such a large increase in grammar size as to be impractical.

An alternative method for making left-corner parsers more robust is to explicitly add predictions for start categories at every point in the input. If every context-independent category is a possible left corner of a start category, this approach will result in the same set of edges in the chart that the limited left-context approach builds, but at the added expense of creating many more predictions. Since increasing the total number of predictions increases parse time, we expect that this technique would be significantly slower than limited left-context checking, although we have not carried out any experiments on this approach.

The technique of precompiling the left-daughter-of table is not unique to this parser, and has appeared in both the GHR parser (Graham, Harrison, and Russo, 1980) and the Core Language Engine parser (Alshawi, 1992).

# INTERLEAVED SEMANTIC PROCESSING

The Gemini system allows either syntax-only parsing or parsing with syntactic and semantic processing fully interleaved. In interleaved processing, whenever a syntax rule successfully creates a new syntactic phrase, corresponding semantic rules are applied to construct possible logical forms for the phrase,[3] the logical forms are checked to verify that they satisfy semantic sortal constraints, and edges for interpretations that pass all constraints are added to the chart. In general, this leads to fewer syntactically distinct analyses being present in the chart (since phrases that have no interpretation satisfying sortal constraints do not produce edges), but semantic ambiguity can lead to a greater total number of semantically distinct edges. As is the case in syntax-only parsing, interleaved processing uses packing to collapse analyses for later processing. Analyses are collapsed if they have the same parent nonterminal, incorporating both syntactic and semantic features, and the same semantic sortal properties.

## Deferred Sortal-Constraint Application

In Gemini, there are two sources of semantic ambiguity to be considered when interleaving syntax and semantics in parsing: semantic rule ambiguity and sortal ambiguity. For every syntactic rule of the form:

Rulename: $A_{syn} \rightarrow B_{syn}, C_{syn}$

there are one or more semantic rules indexed on the same rule name:

Rulename:
$(LF_A, A_{sem}) \rightarrow (LF_B, B_{sem}), (LF_C, C_{sem})$

Here, $LF_A$, $LF_B$ and $LF_C$ are logical form expressions indicating how the logical form $LF_A$ is to be constructed from the logical forms of its children $LF_B$ and $LF_C$, and A, B, and C are category expressions that are unified.

The second source of semantic ambiguity is sortal ambiguity. Every atom in a logical form expression is assigned one or more semantic sorts. For example, in the logical form fragment

```
exists((A;[flight]),
  [and,
    [flight, (A;[flight])];[prop],
    [to, (A;[flight]),
      ('BOSTON';[city])];[prop]];
  [prop]);
[prop]
```

the atoms `exists`, `and`, `flight`, `to` and `'BOSTON'` have sort assignments (sorts are printed as the right-hand side of the ';' operator). Some atoms like 'BOSTON' are assigned atomic sorts like [city], while other atoms like to are assigned more complex sorts, for instance, a function from flights and cities to propositions, represented as ([[flight],[city]],[prop]). Sorts for nonatomic logical form expressions are then constructed recursively from the subexpressions they contain. For instance, the expression [to, (A;[flight]), ('BOSTON';[city])] is assigned the sort [prop] because there is a possible sort assignment for to consistent with the relation to holding between something of sort [flight] and something of sort [city].

---

[3] As a possible optimization, we tried combining the syntactic and semantic rules at compile time. This turned out to be slower than checking all syntactic constraints first, at least for our grammar at the time. We speculate that this is due to unifying multiple variants of the same syntactic pattern against the chart in cases where one syntactic rule has several corresponding semantic rules, and that applying syntactic rules first provides an effective filter for faster matching.

If an atom within a logical form expression has more than one possible sort assignment, then the expression may be ambiguous if the other sorts in the expression do not further constrain it; if a logical form expression associated with a syntactic edge is ambiguous, then new edges are added to the chart for each of the possible semantic readings. This is very common with sort assignments for logical form functors. If all the arguments of the functor have already been found at the point when the functor is first encountered in a logical form expression, then usually only one possible sort assignment for the functor will apply, and the resulting semantic edge will be sortally unambiguous. If the functor is encountered in a phrase where one or more of its arguments have not yet been encountered, such as a verb phrase before it has been combined with its subject, edges for all possible sorts for the missing arguments will be hypothesized, creating local sort ambiguities. As can be seen in Table 2, there is a modest increase in the number of edges created per utterance due to semantic rule ambiguity, but a much more dramatic increase due to sortal ambiguity.

The approach we have taken to deal with this problem is to prevent sortal ambiguity from multiplying out into distinct edges in the chart, by deferring the application of sortal constraints in cases where sortal ambiguities would be created. To implement this approach, we associate with every semantic edge a set (possibly empty) of deferred sort assignments. In order to construct this set for an edge, we create deferred sort assignments for any logical form atoms introduced by the semantic rule or lexical entry that created the edge that have more than one possible sort, given all the information we have at that edge (such as the sorts of the arguments of a functor). For a phrasal edge, we add to this any deferred sort assignments inherited from the daughters of the edge.

Once the set of deferred sorts has been constructed, but before the new edge is added to the chart, the set is analyzed to determine whether it is consistent, and to remove any deferred sort assignments that have become unambiguous because of unifications performed in creating the edge. Since the deferred sort assignments can share logic variables, it is possible that even though each deferred assignment is ambiguous, there is no assignment of sorts that can satisfy all constraints at the same time, in which case the edge is rejected. The incorporation of additional information from sibling nodes can result in a sortal ambiguity becoming resolved when an edge is constructed, in which case the resulting sort assignment is applied and removed from the set of deferred sort assignments. Finally, we check whether the deferred sort assignments, although individually ambiguous, jointly have a unique solution. In this case, that assignment of values is applied, and the set of deferred sort assignments becomes the empty set.

| Type of Processing | Edges/ Utt | Secs/ Edge | Secs/ Utt |
|---|---|---|---|
| Syntax Only | 203 | 0.005 | 0.98 |
| Plus Semantic Rules | 209 | 0.006 | 1.20 |
| Plus Sorts | 357 | 0.011 | 4.04 |
| With Deferred Sorts | 194 | 0.007 | 1.33 |

Table 2: Results of Deferring Sortal Constraints

The effectiveness of this technique is demonstrated by Table 2, which compares the average number of edges per utterance, average parse time per edge, and average parse time per utterance for four different modes of processing: syntax-only parsing, interleaving syntax and semantics without applying sortal constraints, interleaving syntax and semantics while immediately applying sortal constraints, and interleaving syntax and semantics while deferring ambiguous sortal constraints. We can see that the total number of semantic edges is reduced significantly, resulting in a decrease in the total syntax+semantics+sorts time by a factor of 3. Note that despite the addition of semantic rule ambiguity, the total number of edges built during interleaved syntactic and semantic processing is less than the number of edges built using syntax alone, demonstrating that we in fact succeed in using semantic information to prune the syntactic search space.

# IMPROVING ACCURACY IN SPEECH RECOGNITION

One of our prime motivations in designing a parser to find all syntactically well-formed semantically meaningful phrases in a word string was to be able to use it for the robust application of natural-language constraints in speech recognition. Most attempts to apply natural-language constraints in speech recognition have relied on finding a com-plete parse for a recognition hypothesis. Many have worked by simply picking as the preferred hypothesis the string with the highest recognition score that can be completely parsed and interpreted.

It seems virtually impossible, however, to create a natural-language grammar that models spontaneous spoken language accurately enough to avoid introducing more errors than it corrects, if applied in this way. A state-of-the-art natural-language grammar for a problem such as the ARPA ATIS task might fail to find a complete analysis for 10% or more of test utterances. In this case, a substantial recognition error rate would be introduced, because of the correct utterances that would be completely excluded, and it is extremely unlikely that the grammar would result in enough reduction of the recognition errors of a state-of-the-art speech recognizer on other utterances to overcome the errors it introduces.

We have taken a different approach based on the observation that, even when our grammar fails to provide a complete analysis of an utterance, it is usually possible to find a small number of semantically meaningful phrases that span the utterance. We therefore use our parser to find the minimal number of semantically meaningful phrases needed to span a recognition hypothesis and to compute a natural-language score for the hypothesis based on this number. Having a parser that finds all syntactically well-formed semantically meaningful phrases is an obvious prerequisite to taking such an approach.

We have applied this idea in a system combining Gemini with SRI's DECIPHER$^{TM}$ speech recognizer (Murveit et al., 1993), which was tested in the December 1993 ARPA ATIS benchmark evaluation (Pallet et al., 1994). The following example from the evaluation test set illustrates the basic approach:

hypothesis: [*list flights*][*of fare code*][*a*][*q*]
reference: [*list flights*][*of fare code of q*]

These two word strings represent the recognizer's first hypothesis for the utterance and the reference transcription of the utterance, each bracketed according to the best analysis that Gemini was able to find as a sequence of semantically meaningful phrases. Because of a missing sortal possibility, Gemini did not allow the preposition *of* to relate a noun phrase headed by *flights* to a noun phrase headed by *fare code*, so it was not possible to find a single complete analysis for either word string. Gemini was, however, able to find a single phrase spanning *of fare code of q*, but required three phrases to span *of fare code a q*, so it still strongly preferred the reference transcription of the utterance over the recognizer's first hypothesis.

The integration of Gemini and DECIPHER was implemented by combining a Gemini score with the recognition score for each of the recognizer's $N$-top hypotheses and selecting the hypothesis with the best overall score.[4] The Gemini score was computed as a somewhat ad hoc combination of the number of phrases needed to cover the hypothesis, a bonus if the hypothesis could be analyzed as a single sentence (as opposed to any other single grammatical phrase), and penalties for using certain "dispreferred" grammar rules. This score was then scaled by an empirically optimized parameter and added to the recognition score.

We carried out a detailed analysis of the preliminary results of the December 1993 ARPA ATIS benchmark evaluation to determine the effect of incorporating natural-language information into recognition in this way. Overall, the word error rate improved from 6.0% to 5.7% (5.0% improvement), and the utterance error rate improved from 29.6% to 27.8% (6.1% improvement). These improvements, while modest, were measured to be statistically significant at the 95% confidence level according to the matched-pair sentence segment (word error) test and the McNemar (sentence error) test.

In more detail, the first hypothesis of the recognizer was correct for 704 of 995 utterances for which the natural-language grammar was used. Of these, the natural-language grammar failed to find complete analysis for 62. The combined system nevertheless chose the correct hypothesis in 57 of these cases; thus, only 5 correct hypotheses were lost due to lack of grammar coverage. On the other hand, use of the natural-language grammar resulted in correcting 22 incorrect recognizer first hypotheses. Moreover, 4 of these were not completely analyzable by the natural-language grammar, but were chosen because they received a better analysis as a sequence of phrases than the first hypothesis of the recognizer.

We also analyzed which of the natural-language factors incorporated in the Gemini score were responsible for the corrections and errors relative to the performance of the recognizer alone. For the 22 utterances that were corrected, in 18 cases the correction was due to the preference for fewer fragments, in 3 cases the correction was due to the preference for complete sentences, and in only one case did the correction result from a grammar rule preference. Of the 5 utterance errors introduced by Gemini, 3 turned out to be cases in which the reference transcription was incorrect and the hypothesis selected by Gemini was actually correct, one was due to inadequate grammatical coverage resulting in a larger number of fragments for the correct hypothesis, and one was due to a grammatical rule preference. We concluded from this that the preference for fewer fragments is clearly useful and the preference for complete sentences seems to be somewhat useful, but there is no evidence that the current system of rule preferences is of any benefit in speech recognition. A more systematic approach to rule preferences, such as one based on a statistical grammar, may be of more benefit, however.

---

[4]The value of $N$ was variable, but sufficiently large (typically hundreds) that a limit on $N$ was never a factor in which hypothesis was chosen.

## CONCLUSIONS

We have described an efficient parser that operates bottom-up to produce syntactic and semantic structures fully interleaved. Two techniques combine to reduce the total ambiguity represented in the chart. Limited left-context constraints reduce local syntactic ambiguity, and deferred sortal-constraint application reduces local semantic ambiguity. We have expermentally evaluated these techniques, and shown order-of-magnitude reductions in both number of chart edges and total parsing time. The robust processing capabilities of the parser have also been shown to be able to provide a small but significant increase in the accuracy of a speech recognizer.

## ACKNOWLEDGMENTS

We would like to thank Mark Gawron for helpful comments on earlier drafts, and the SRI speech group, particularly Harry Bratt, for help performing the speech recognition experiments.